\documentclass{article}

\usepackage{microtype}
\usepackage{graphicx}
\usepackage{subfigure}
\usepackage{booktabs} 

\usepackage{hyperref}

\usepackage[accepted]{icml2021}

\usepackage{bibentry}

\usepackage{amsmath,amsfonts,amssymb,amsthm,dsfont}

\def\R{\mathbb{R}}

\def\onehot{\mathrm{onehot}}
\def\cost{\mathrm{cost}}
\def\LogSumExp{\mathrm{LogSumExp}}

\def\our{\mbox{ProPaLL}}

\usepackage{booktabs}       
\usepackage{url}

\usepackage{makecell}
\usepackage{caption}
\usepackage{lscape}
\usepackage{multirow}

\usepackage[capitalize,nosort]{cleveref}
\Crefname{section}{Section}{Sections}
\crefname{section}{Sec.}{Secs.}
\Crefname{table}{Table}{Tables}
\crefname{table}{Tab.}{Tabs.}
\Crefname{appsec}{Appendix}{Appendices}
\crefname{appsec}{App.}{Apps.}

\usepackage[toc,page]{appendix}

\begin{document}

\twocolumn[
\icmltitle{ProPaLL: Probabilistic Partial Label Learning}


\begin{icmlauthorlist}
\icmlauthor{\L{}ukasz Struski}{to}
\icmlauthor{Jacek Tabor}{to}
\icmlauthor{Bartosz Zieli\'nski}{to}
\end{icmlauthorlist}

\icmlaffiliation{to}{Faculty of Mathematics and Computer Science, 
Jagiellonian University, Krak\'ow, Poland}

\icmlcorrespondingauthor{\L{}ukasz Struski}{lukasz.struski@uj.edu.pl}
\icmlcorrespondingauthor{Jacek Tabor}{jacek.tabor@uj.edu.pl}
\icmlcorrespondingauthor{Bartosz Zieli\'nski}{bartosz.zielinski@uj.edu.pl}


\vskip 0.3in
]



\printAffiliationsAndNotice{}  

\begin{abstract}
Partial label learning is a type of weakly supervised learning, where each training instance corresponds to a set of candidate labels, among which only one is true. In this paper, we introduce ProPaLL, a novel probabilistic approach to this problem, which has at least three advantages compared to the existing approaches: it simplifies the training process, improves performance, and can be applied to any deep architecture. Experiments conducted on artificial and real-world datasets indicate that ProPaLL outperforms the existing approaches.
\end{abstract}

\section{Introduction}

Deep neural networks achieve excellent performance in many real applications. However, they require a large-scale training set with correctly labeled samples. Obtaining such a set is time-consuming, expensive, and, in some domains, impossible due to discrepancies between labeling experts~\cite{armato2011lung}. In consequence, many contemporary datasets are weakly labeled, forcing researchers to propose adequate learning strategies within the Weakly Supervised Learning paradigm~\cite{zhou2018brief}.

The most common problem setting in supervised learning assumes that the class to which the training data belongs is provided as a label, namely an ordinary label. However, in some cases, each training instance is associated with a set of candidate labels, among which exactly one is true. This setting, known as Partial Label Learning (PLL)~\cite{jin2002learning}, occurs in many real-world tasks, such as image annotation~\cite{chen2017learning}, web mining~\cite{luo2010learning}, multimedia content analysis~\cite{zeng2013learning,chen2017learning}, and ecoinformatics~\cite{liu2012conditional,tang2017confidence}.
Moreover, in the special case of PLL, called Complementary Label Learning (CLL)~\cite{ishida2017learning}, the number of candidate labels is only one less than the number of classes, making this task even more challenging..

The standard approaches to partial and complementary label learning are incompatible with high-efficient stochastic optimization and cannot handle large-scale datasets~\cite{liu2012conditional}. On the other hand, more contemporary methods use deep networks with stochastic optimizers as a backbone, but they are restricted, e.g. to some specific architectures~\cite{yao2020deep}. Moreover, they commonly rely on modeling the relationship between the complementary and ground-truth labels for each training instance. For this purpose, they often use the iterative Expectation-Maximization (EM) technique, where M-step learns a predictor while E-step increases weights of more possible labels every few training epochs~\cite{tang2017confidence,feng2019partial,lv2020progressive}, resulting in a suboptimal solution. Therefore, partial and complementary label learning still brings many challenges.

\begin{figure}[t]
    \centering
    \includegraphics[width=\columnwidth]{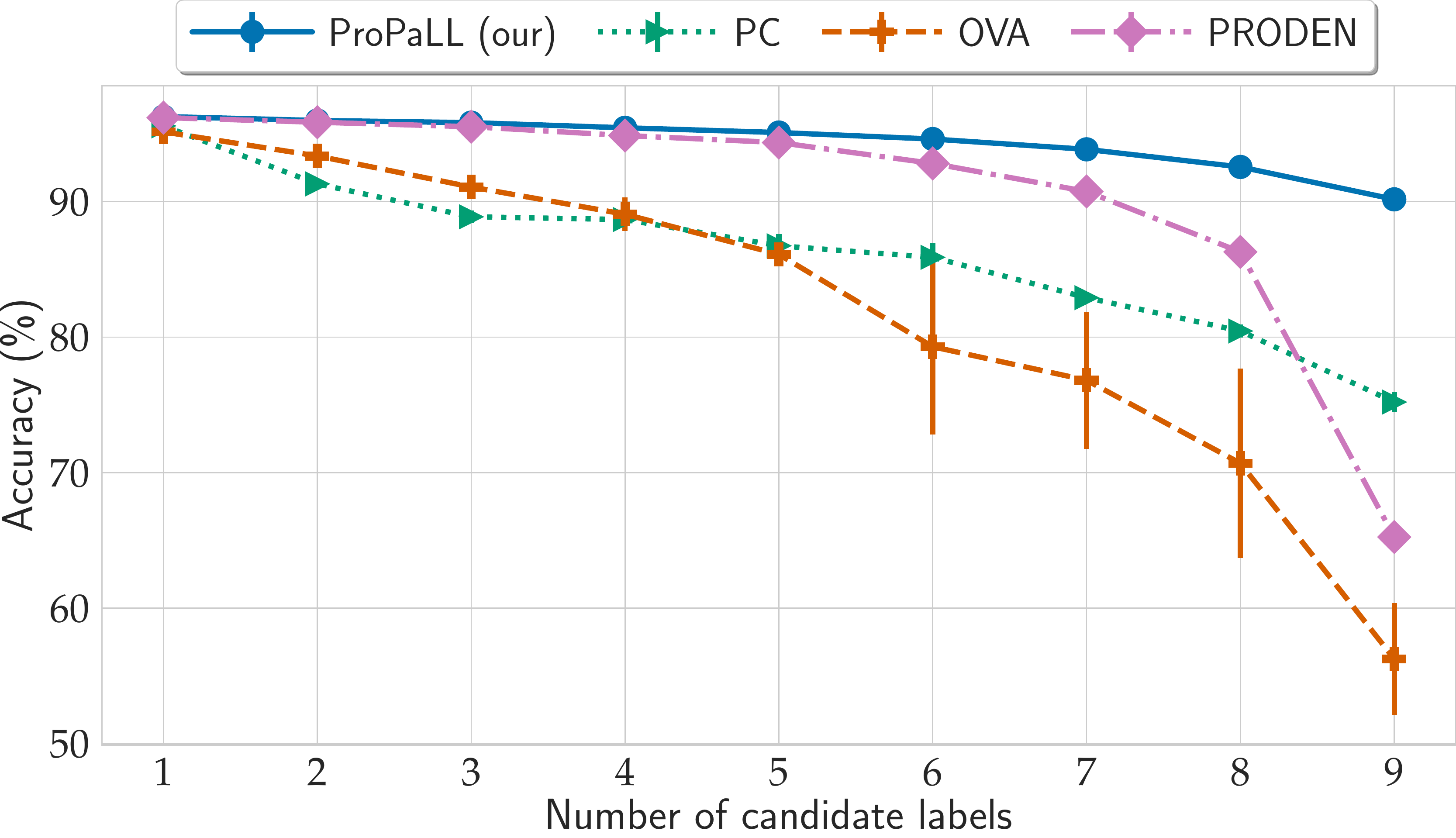}
    \caption{Accuracy of our ProPaLL and the popular PLL approaches for the MNIST dataset depending on the number of candidate labels. While the performance of baseline PLL methods drops rapidly for more complicated setups, our method maintains $90\%$ accuracy even for $9$ candidate labels.}
    \label{fig:acc_vs_no}
\end{figure}

\begin{figure*}[ht]
    \centering
    \includegraphics[width=\textwidth]{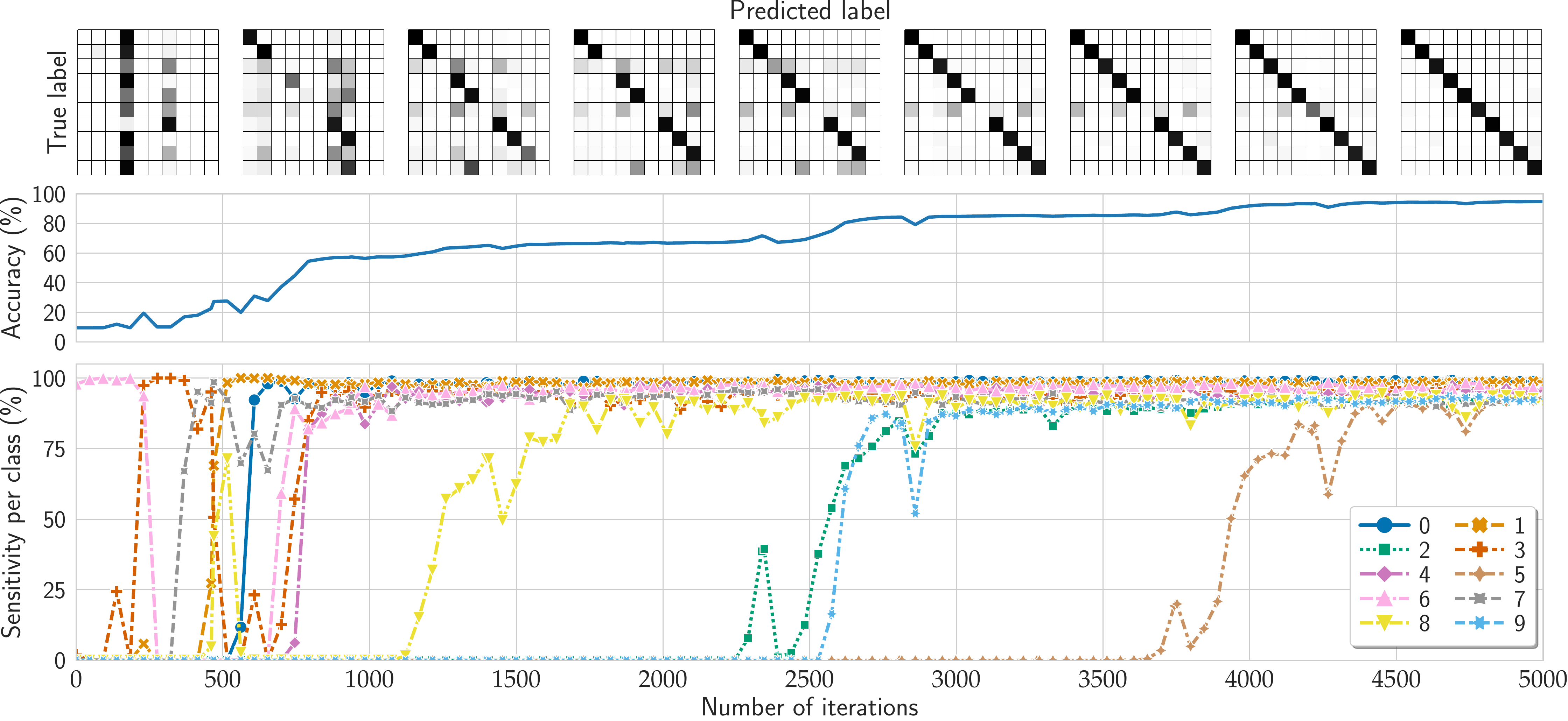}
    \caption{Detailed analysis of ProPaLL training indicates that it gradually learns consecutive classes instead of progressively increasing sensitivity for all of them. In this image, the middle part presents model accuracy for all classes in time, the bottom part shows the sensitivity of particular classes also in time, while the top part shows confusion matrices in time points corresponding to their location. A model starts with learning classes 1, 3, 4, 6, and 7 (before iteration 1000). Then, around iteration 1500, it learns class 8, while classes 2 and 9 are trained around iteration 2700. Finally, it learns class 5 near iteration 4300.}
    \label{fig:jump_acc}
\end{figure*}

In this paper, we introduce \our{}\footnote{Our code with $\our$ is available on the GitHub repository \url{https://github.com/anonymised}.} (abbr. from Probabilistic Partial Label Learning), a novel probabilistic approach to partial label learning problems to address the abovementioned issues. For this purpose, we modify Binary Cross Entropy (BCE) loss to maximize the probability that at least one index corresponding to candidate labels equals $1$ and all indexes complementary to candidate labels equal $0$. 
We show that \our{} can be potentially applied to any deep architecture and target tasks (such as classification, detection, or segmentation) in both partial and complementary settings without adjusting any additional hyperparameters. The experiments conducted on four artificial and five real-world datasets indicate that ProPaLL outperforms existing PLL methods, especially if the number of candidate labels increases (see~\Cref{fig:acc_vs_no}). Moreover, in contrast to existing methods, it trains faster (see~\Cref{fig:relativeTime}) and gradually learns consecutive classes (see~\Cref{fig:jump_acc}).

Our contributions can be therefore summarized as follows:
\begin{itemize}
    \item We introduce \our{}, a novel probabilistic approach that can be used in both partial and complementary label learning.
    \item Our approach requires only loss function modification and, as such, can be applied to any deep architecture without adjusting any additional hyperparameters.
    \item Despite its simplicity, the performance and efficiency of \our{} outperforms the existing more complex methods, both for artificial and real-world datasets.
\end{itemize}

\section{Related works}

\paragraph{Partial label learning.}
Existing PLL approaches often aim to fit widely-used learning techniques to disambiguate partial label data. For example, \cite{jin2002learning} and \cite{liu2012conditional} maximize the likelihood defined over the candidate label set. \cite{hullermeier2006learning} and~\cite{zhang2015solving} determine the class label of an unseen instance via voting among the candidate labels of its k-nearest neighbors. \cite{nguyen2008classification} and~\cite{yu2016maximum} maximize the classification margins between candidate and non-candidate labels. While boosting techniques~\cite{tang2017confidence} update the weights and confidence over the candidate labels in each boosting round.
Disambiguation strategies ignore considering the generalized label distribution, in contrast to disambiguation-free strategies~\cite{xu2019partial,zhang2016partial}, which learn a multi-class predictive model by fitting a regularized multi-output regressor with the generalized label distributions (recovered, e.g., by leveraging the topological information of the feature space).
The abovementioned methods have low efficiency and cannot handle large-scale datasets. That is why recent approaches move toward deep networks, employing various regularizers~\cite{yao2020deep} or risk estimators~\cite{feng2020provably,lv2020progressive}. Moreover, most recent approaches~\cite{xu2021instance} consider instance-dependent (feature-dependent) candidate labels.

\paragraph{Complementary label learning.}
CLL was first proposed in~\cite{ishida2017learning}, where modified one-versus-all (OVA) and pairwise-comparison (PC) losses are proposed.
\cite{yu2018learning} introduce a different formulation for complementary labels by employing the forward loss correction technique~\cite{patrini2017making} but limit the loss function to softmax cross-entropy loss.
This limitation is overcome by~\cite{ishida2019complementary}, who propose a new unbiased risk estimator, allowing usage of any loss (convex, non-convex) and any model (parametric, non-parametric).
Moreover, one of the most recent approaches generalizes OVA and PC loss functions for more than one complementary label for each training instance~\cite{katsura2020bridging}. \cite{xu2020generative} applies conditional generative adversarial networks~\cite{goodfellow2014generative}. Finally, \cite{gao2021discriminative} adopts weighted loss to the empirical risk to maximize the predictive gap between the potential ground-truth and complementary labels.
Our method introduces a novel loss function that overpasses the performance of existing approaches in PLL and CLL setups.

\section{Probabilistic Partial Label Learning}

In this section, we introduce the loss function used by \our{}. However, since it bases on Binary Cross Entropy (BCE), we first establish notation and derive the BCE formula in the standard multi-label classification problem.

\paragraph{Notations.}
Let us consider the classification task for $k$ classes with the product $\{0,1\}^k$ at the output. Moreover, let $p_i$ denotes the probability of obtaining $1$ at $i$-th output coordinate. Then, the probability of drawing a binary sequence $b=(b_1,\ldots,b_k) \in \{0,1\}^k$ is defined as
$$
P(b)=\prod_{i=1}^k p_i^{b_i} (1-p_i)^{1-b_i}.
$$
Consequently, the probability of predicting class $c$ (i.e. obtaining $1$ at $c$-th coordinate and zero for all other coordinates) is given by
\begin{equation} \label{eq:bce}
P(\onehot(c))=p_c \cdot \prod_{j \neq c} (1-p_j).
\end{equation}

\paragraph{BCE loss for ordinary labels.}
In the most common problem setting of supervised learning, each point $x$ from the dataset $X \subset \R^d$ is associated with a class $c \in \{1,\ldots,k\}$, and we aim to train a network $F_\theta:\R^d \to \R^k$
$$
F_\theta(x)=(r_1(x),\ldots,r_{k}(x)) \in \R^k,
$$
where $\theta$ are the network weights and logits $r_i(x)$ denotes the network output before the sigmoid $\sigma(r)=1/(1+\exp(-r))$. The output after sigmoid corresponds to the probability of predicting class $i$
$$
p_i(x)=\sigma(r_i(x)),
$$
and the opposite probability, i.e. the probability of predicting class other than $i$ equals
$$
1-\sigma(r_i(x))=\sigma(-r_i(x)).
$$
Thus, by~\eqref{eq:bce} probability for point $x$ and class $c$ is defined as
$$
P(\onehot(c))=p_{c}(x) \cdot \prod_{j \neq c}(1-p_j(x)),
$$
and consequently the BCE cost is given as the minus log-likelihood:
\begin{equation} \label{eq:BCE2}
\cost(x)=-\log(p_{c}(x))-\sum_{j \neq c}\log(1-p_j(x)).
\end{equation}
Since
$$
-\log \sigma(r)=-\log \tfrac{1}{1+\exp(-r)}=
\LogSumExp(0,-r)
$$
and
$$
-\log(1-\sigma(r))=-\log \sigma(-r)=
\LogSumExp(0,r), 
$$
we can rewrite the cost function in terms of the logits by
$$
\begin{array}{l}
\cost(x)=\\[1ex]
\LogSumExp(0,-r_{c}(x))+\sum\limits_{j \neq c} \LogSumExp(0,r_j(x)).
\end{array}
$$
In the evaluation, we associate $x$ with the label obtaining the greatest probability.

\paragraph{BCE adaptation for partial label learning.}
In the partial label learning setting, each point $x \in X$ is associated with a candidate label set
$
S \subset \{1,\ldots,k\},
$
where exactly one $c \in S$ is true. Therefore, to adapt BCE into this setting, we introduce $\mathcal{S} \subset \{0,1\}^k$, the set with at least one $1$ in the set $S$, and $0$ outside $S$
$$
\mathcal{S}=\{b \in \{0,1\}^k : \sum_{i \in S} b_i \geq 1, \sum_{j \not\in S}b_j=0\},
$$
and we aim to maximize the probability of $\mathcal{S}$
\begin{equation} \label{eq:11}
P(\mathcal{S})=\big(1-\prod_{i \in S}(1-p_i(x)) \big) \cdot  \prod_{ j\not\in S} (1-p_j(x)).
\end{equation}
The first part corresponds to the opposite probability of occurring only $0$ in $S$ and the second part to the probability of occurring only $0$ in labels complementary to $S$. It can seem counterintuitive that we do not penalize the model for predicting more than one label from $S$, but thanks to that, the network can easily switch from one class to another when it learns how to classify a given point.

\begin{table}[t]\fontsize{9.8}{10.8}\selectfont 
\centering
\begin{tabular}{l@{\;}r@{\;\;}r@{\;\;}r@{\;\;}r@{\;\;}r@{}} 
\toprule
\multirow{2}{*}{Datasets} & \multirow{2}{*}{\#Train} & \multirow{2}{*}{\#Test} & \multirow{2}{*}{\#Feats} & \multicolumn{2}{c}{Labels} \\
\cmidrule{5-6} 
&  &  & & \#Class & Avg. \#S \\
\midrule
Lost & $898$ & $224$ & $108$ & $16$ & $2.23$ \\ 
Soccer Player & $13978$ & $3494$ & $279$ & $171$ & $2.09$ \\ 
Yahoo! News & $18393$ & $4598$ & $163$ & $219$ & $1.91$ \\ 
MSRCv2 & $1406$ & $352$ & $48$ & $23$ & $3.16$ \\ 
BirdSong & $3998$ & $1000$ & $38$ & $13$ & $2.18$ \\ 
 \bottomrule
 \end{tabular}
\caption{Description of the real-world partial label datasets.}
\label{tab.real_datasets_info}
\end{table}

\paragraph{\our{} cost function.}
Based on the previous paragraph, we finally define the cost function of \our{} as the minus log-likelihood of probability defined in \Cref{eq:11}:
\begin{equation} \label{eq:pll}
\begin{array}{l}
\cost(x)=-\log P(\mathcal{S}) \\[0.5ex]
=-\log \big(1-\prod \limits_{i \in S} (1-p_i(x)) \big)-\sum \limits_{j \not\in S} \log(1-p_j(x)),
\end{array}
\end{equation}
and in terms of the logits, as
$$
\begin{array}{l}
\cost(x) = \\[1ex]
-\log \big(1-\exp(-\sum \limits_{i \in S} \LogSumExp(0,r_i(x))
) \big) \\[0.5ex]
+\sum \limits_{j \not\in S} \mathrm{LogSumExp}(0,r_j(x)).
\end{array}
$$

Please note that when $S$ is one-element set $\{c\}$, it reduces to \eqref{eq:BCE2}. Moreover, in Appendices, we provide details on the implementation of \our{} loss function.

\begin{table}[t]\fontsize{8}{10}\selectfont
\centering
 \begin{tabular}{l@{}c@{\,}c@{\,}c@{\,}c@{}} 
 \toprule
 \multirow{2}{*}{Methods} & \multicolumn{4}{c}{Datasets} \\
 \cmidrule{2-5}
  & MNIST & Kuzushiji & Fashion & CIFAR-10 \\ 
 \midrule
 \our{} & \boldmath{$98.73(0.05)$} & \boldmath{$93.50(0.17)$} & \boldmath{$90.40(0.06)$} & \boldmath{$84.32(0.37)$} \\ 
 VALEN & $97.93\,(0.05)$ & $88.76\,(0.26)$ & $88.98\,(0.16)$ & $81.53\,(0.53)$ \\
 PRODEN & $97.97\,(0.03)$ & $88.55\,(0.10)$ & $88.94\,(0.12)$ & $81.61\,(1.52)$ \\
 RC & $97.86\,(0.03)$ & $86.65\,(0.10)$ & $88.59\,(0.08)$ & $81.30\,(1.30)$ \\
 CC & $97.73\,(0.02)$ & $87.99\,(0.03)$ & $88.93\,(0.06)$ & $80.17\,(1.09)$ \\
 D$^2$CNN & $95.12\,(0.16)$ & $84.03\,(0.78)$ & $80.42\,(0.21)$ & $75.11\,(0.11)$ \\
 GA & $96.29\,(0.19)$ & $82.36\,(0.98)$ & $81.81\,(0.99)$ & $60.14\,(1.35)$ \\
 \bottomrule
 \end{tabular}
 \caption{Test accuracy (mean\,(std)\%) on artificial datasets corrupted by the uniform generating procedure. The highest accuracy is bolded.
}
\label{tab.bemchmark_random}
\end{table}

\section{Experimental Setup}

In this section, we describe datasets used in the evaluation and the baseline methods to which we compare our ProPaLL approach.

\paragraph{Datasets.}
First of all, we consider four artificial datasets, created based on MNIST~\cite{lecun1998gradient}, Fashion-MNIST~\cite{xiao2017fashion}, Kuzushiji-MNIST~\cite{clanuwat2018deep}, and CIFAR-10~\cite{krizhevsky2009learning}. All these datasets contain $60000$ images split into $50000$ training and $10000$ testing sets. The CIFAR-10 has $32\times32$ RGB images, while the remaining datasets have $28\times28$  grayscale images. Moreover, in the original version of those datasets, each point corresponds to one of the $10$ classes. Hence, to modify them to a partial label learning setting, we corrupt each training data point by associating it with a candidate label set containing the original label and a few other labels sampled uniformly or using an instance-dependent process~\cite{xu2021instance}.

In addition, we use five real-world partial label datasets: Lost~\cite{cour2011learning}, Soccer Player~\cite{zeng2013learning} and Yahoo!News~\cite{guillaumin2010multiple} for automatic face naming from images or videos, MSRCv2~\cite{liu2012conditional} for object classification, and BirdSong~\cite{briggs2012rank} for bird song classification. \Cref{tab.real_datasets_info} shows the statistics on these datasets, including the number of elements in the training and test set, the number of features, the number of classes, and the average number of candidate labels.

\paragraph{PLL baseline methods.}
We compare our approach to six state-of-the-art partial label learning methods:
\begin{itemize}
\item VALEN~\cite{xu2021instance} recovers the label distribution and trains the predictive model iteratively in each epoch.
\item PRODEN~\cite{lv2020progressive} updates the model and identification of true labels in a seamless manner.
\item RC~\cite{feng2020provably} uses the reweighting strategy to converge the true risk minimizer.
\item CC~\cite{feng2020provably} uses a transition matrix to form an empirical risk estimator.
\item D$^2$CNN~\cite{yao2020deep} uses an entropy-based regularizer to maximize the margin between the potentially correct label and the unlikely ones.
\item GA~\cite{ishida2019complementary} is an unbiased risk estimator approach.
\end{itemize}

Moreover, in the case of real-world datasets, we additionally compare our method to five classical approaches:
\begin{itemize}
\item CLPL~\cite{cour2011learning} uses averaging-based disambiguation.
\item PL-SVM~\cite{nguyen2008classification} uses identification-based disambiguation.
\item PL-KNN~\cite{hullermeier2006learning} uses k-nearest neighbor weighted voting.
\item IPAL~\cite{zhang2017disambiguation} is a non-parametric method that applies the label propagation strategy to iteratively update the confidence of each candidate label.
\item PLLE~\cite{xu2019partial} estimates the generalized description degree of each class label value via graph Laplacian.
\end{itemize}

Like in~\cite{lv2020progressive,xu2021instance}, we employ a $5$-layer perceptron for MNIST, Fashion-MNIST, Kuzushiji-MNIST datasets, and ResNet-32~\cite{he2016deep} for CIFAR-10. Moreover, we train each model for $500$ epochs and use the stochastic gradient descent with momentum $0.9$ to optimize models. Hyperparameters are the same as in the original papers. Details are provided in Appendices.

\begin{table}[t]\fontsize{8}{10}\selectfont
\centering
 \begin{tabular}{l@{}c@{\,}c@{\,}c@{\,}c@{}} 
 \toprule
 \multirow{2}{*}{Methods} & \multicolumn{4}{c}{Datasets} \\
 \cmidrule{2-5}
  & MNIST & Kuzushiji & Fashion & CIFAR-10 \\ 
 \midrule
\our{} & \boldmath{$97.89(0.08)$} & \boldmath{$86.26(0.32)$} & $85.55\,(0.33)$ & $75.75\,(0.63)$ \\ 
 VALEN & $97.85\,(0.05)$ & $86.19\,(0.14)$ & \boldmath{$86.17(0.19)$} & \boldmath{$80.38(0.52)$} \\
 PRODEN & $97.69\,(0.04)$ & $85.71\,(0.12)$ & $85.54\,(0.09)$ & $79.80\,(0.28)$ \\
 RC & $97.60\,(0.05)$ & $84.86\,(0.11)$ & $85.51\,(0.10)$ & $79.46\,(0.25)$ \\
 CC & $97.44\,(0.03)$ & $82.67\,(1.82)$ & $85.19\,(0.04)$ & $78.98\,(0.60)$ \\
 D$^2$CNN & $94.63\,(0.16)$ & $83.03\,(0.78)$ & $82.42\,(0.21)$ & $73.11\,(0.11)$ \\
 GA & $95.25\,(0.07)$ & $82.45\,(0.63)$ & $80.41\,(0.24)$ & $77.57\,(0.76)$ \\
 \bottomrule
 \end{tabular}
 \caption{Test accuracy (mean\,(std)\%) on artificial datasets corrupted by the instance-depending generating procedure~\cite{xu2021instance}. The highest accuracy is bolded.
}
\label{tab.bemchmark_instance_dependent}
\end{table}

\begin{figure}[t]
    \centering
    \includegraphics[width=\columnwidth]{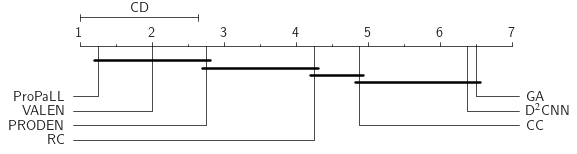}
    \caption{Critical difference diagrams comparing results of the considered methods shown in~\Cref{tab.bemchmark_random,tab.bemchmark_instance_dependent} (smaller is better). ProPaLL performs significantly better than all baseline methods except VALEN and PRODEN.}
    \label{fig:cd_pll}
\end{figure}

\begin{table*}[t]\fontsize{9.8}{10.8}\selectfont 
\centering
 \begin{tabular}{l@{\;\;}c@{\;\;}c@{\;\;}c@{\;\;}c@{\;}c@{}} 
 \toprule
 \multirow{2}{*}{Methods} & \multicolumn{5}{c}{Datasets} \\
 \cmidrule{2-6}
 & Lost & MSRCv2 & BirdSong & Soccer Player & Yahoo!News \\ 
 \midrule
 \our{} (our) & \boldmath{$78.78\,(0.22)$} & $42.05\,(0.58)$ & $70.49\,(0.56)$ & \boldmath{$56.73\,(0.27)$} & \boldmath{$68.06\,(0.08)$} \\
 VALEN & $70.28\,(2.29)$ & $47.61\,(1.79)$ & \boldmath{$72.02\,(0.37)$} & $55.90\,(0.58)$ & $67.52\,(0.19)$ \\
 PRODEN & $68.62\,(4.86)$ & $44.47\,(2.33)$ & $71.68\,(0.83)$ & $54.40\,(0.85)$ & $67.12\,(0.97)$ \\
 RC & $68.89\,(5.02)$ & $44.59\,(2.65)$ & $71.56\,(0.88)$ & $54.23\,(0.89)$ & $67.04\,(0.88)$ \\
 CC & $62.21\,(1.77)$ & $47.49\,(2.31)$ & $68.42\,(0.99)$ & $53.50\,(0.96)$ & $61.92\,(0.96)$ \\
 D$^2$CNN & $68.56\,(6.68)$ & $43.27\,(2.98)$ & $65.48\,(2.57)$ & $48.16\,(0.62)$ & $52.46\,(1.71)$ \\
 GA & $50.21\,(3.62)$ & $30.91\,(4.31)$ & $34.57\,(3.41)$ & $50.65\,(0.94)$ & $45.72\,(1.75)$ \\
 CLPL & $74.15\,(3.03)$ & $44.47\,(2.58)$ & $65.76\,(1.19)$ & $50.01\,(1.03)$ & $53.25\,(1.12)$ \\
 PL-SVM & $71.56\,(2.71)$ & $38.25\,(3.89)$ & $50.66\,(4.23)$ & $36.39\,(1.03)$ & $51.24\,(0.72)$ \\
 PL-KNN & $33.87\,(2.48)$ & $43.28\,(2.35)$ & $64.34\,(0.75)$ & $49.24\,(1.23)$ & $40.38\,(0.37)$ \\
 IPAL & $72.10\,(2.75)$ & \boldmath{$52.96\,(1.36)$} & $70.32\,(0.91)$ & $54.41\,(0.68)$ & $66.04\,(0.85)$ \\
 PLLE & $72.55\,(3.55)$ & $47.54\,(1.96)$ & $70.63\,(1.24)$ & $53.38\,(1.03)$ & $59.45\,(0.43)$ \\
 \bottomrule
 \end{tabular}
\caption{Test accuracy (mean\,(std)\%) on real-world datasets. The highest accuracy is bolded.
}
\label{tab.real_datasets}
\end{table*}

\begin{figure}[t]
    \centering
    \includegraphics[width=\columnwidth]{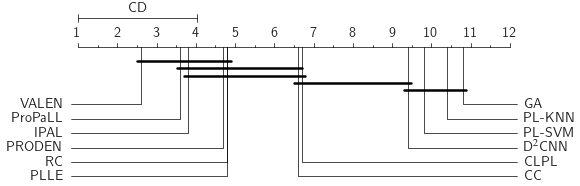}
    \caption{Critical difference diagrams comparing results of the considered methods shown in~\Cref{tab.real_datasets} (smaller is better).}
    \label{fig:cd_pll_realdata}
\end{figure}

\paragraph{CLL baseline methods.}
In these scenario, we used the typical datasets described in the previous paragraph, namely MNIST, Fashion-MNIST, Kuzushiji-MNIST, CIFAR-10. We compare our method to few state-of-the-art aproches:
\begin{itemize}
\item PC and OVA~\cite{katsura2020bridging} use pairwise classification or one-versus-all in training.
\item L-W and L-UW~\cite{gao2021discriminative} are based on the weighted and unweighted losses, respectively.
\item Forward~\cite{yu2018learning} applies forward loss correction.
\item NN and GA~\cite{ishida2019complementary} improve the risk estimator by a non-negative correction and gradient ascent trick, respectively.
\end{itemize}
We also use GA and PRODEN methods presented in the previous paragraph.

Similarly to~\cite{katsura2020bridging} and~\cite{gao2021discriminative}, we use a 2-layer perceptron for the MNIST, Fashion-MNIST, and Kuzushiji-MNIST datasets, and DenseNet~\cite{huang2017densely} for the CIFAR-10. We train each model for $300$ epochs and apply the Adam optimizer and stochastic gradient descent for the MLP and DenseNet, respectively. Again, hyperparameters are the same as in the original papers.

\paragraph{ProPaLL setup.}
We run the efficient implementation of the loss function~\eqref{eq:pll} described in the Appendices, with the same number of epochs and optimizers as for the baseline methods. Moreover, we add a Gumbel random noise to the logits ($r$)
\begin{equation} \label{eq:GS}
r_i = r_i + \lambda \cdot (U_i - V_i)
\end{equation}
where $U_i,V_i$ are drawn independently from the Gumbels distribution, and $\lambda$ is set to $1$ for $80\%$ of training and decreasing linearly to zero afterwards. This way, the model obtains more relevant global minima~\cite{struski2021song,wang2021noisy}. The derivation of the~\Cref{eq:GS} is provided in the Appendices. All experiments were implemented using PyTorch\footnote{\url{https://pytorch.org}} and run on NVIDIA GeForce RTX $3080$.

\section{Results and Discussion}

This section compares our method with the baseline approaches in PLL and CLL settings. Moreover, we present the intriguing behavior of ProPaLL during training and the positive influence of Gumbel noise.

\paragraph{Results on PLL setting.}
In~\Cref{tab.bemchmark_random,tab.bemchmark_instance_dependent}, we report test accuracy for artificial databases corrupted by the uniform and instance-dependent generating procedure, respectively\footnote{We reran most of the experiments from~\cite{xu2021instance} and obtained similar results. Therefore, we decided to present the original ones.}. Our method overpasses existing methods for all databases in the case of uniformly corrupted data. Moreover, in the instance-dependent procedure, it is better for MNIST and Kuzushiji-MNIST. The statistical differences are drawn at Critical Difference (CD) diagrams~\cite{demvsar2006statistical} in~\Cref{fig:cd_pll}, generated using all the results shown in~\Cref{tab.bemchmark_random,tab.bemchmark_instance_dependent}. One can observe that \our{}, VALEN, and PRODEN are statistically similar with a preference for our approach.

Moreover, in~\Cref{tab.real_datasets}, we report the results on real-world datasets where our approach achieves the best performance for three out of five databases. Most probably, the weak results for MSRCv2 and BirdSong are caused by shallow architectures and the small scale of those datasets (see~\Cref{tab.real_datasets_info}). In this case, the CD diagram (see~\Cref{fig:cd_pll_realdata}) shows that our method is statistically similar to the best-performing method.

\begin{table*}[h]\fontsize{8}{10}\selectfont
\centering
\begin{tabular}{@{}l@{}l@{}c@{\,}c@{\,}c@{\,}c@{\,}c@{\,}c@{\,}c@{\,}c@{\,}c@{}}
\toprule
& \multirow{2}{*}{Methods} & \multicolumn{9}{c}{Number of candidate labels} \\
\cmidrule{3-11}
& & 1 & 2 & 3 & 4 & 5 & 6 & 7 & 8 & 9 \\
\midrule
\multirow{4}{*}{\rotatebox[origin=c]{90}{MNIST}} & \our{} & \boldmath{$96.28(0.08)$} & \boldmath{$95.99(0.07)$} & \boldmath{$95.83(0.28)$} & \boldmath{$95.45(0.22)$} & \boldmath{$95.10(0.11)$} & \boldmath{$94.62(0.08)$} & \boldmath{$93.86(0.22)$} & \boldmath{$92.56(0.27)$} & \boldmath{$90.17(0.24)$} \\
& PRODEN & $96.20\,(0.05)$ & $95.86\,(0.30)$ & $95.53\,(0.20)$ & $94.88\,(0.20)$ & $94.36\,(0.33)$ & $92.82\,(0.23)$ & $90.75\,(0.16)$ & $86.29\,(0.15)$ & $65.25\,(0.57)$ \\
& PC & $95.59\,(0.22)$ & $91.31\,(0.19)$ & $88.88\,(0.46)$ & $88.68\,(0.31)$ & $86.74\,(0.87)$ & $85.89\,(1.05)$ & $82.91\,(0.31)$ & $80.45\,(0.47)$ & $75.21\,(0.74)$ \\
& OVA & $95.15\,(0.21)$ & $93.37\,(0.28)$ & $91.08\,(0.03)$ & $89.07\,(1.23)$ & $86.11\,(0.64)$ & $79.29\,(6.48)$ & $76.82\,(5.06)$ & $70.69\,(6.99)$ & $56.25\,(4.13)$ \\

\midrule

\multirow{4}{*}{\rotatebox[origin=c]{90}{ Fashion }} & \our{} & \boldmath{$88.73(0.25)$} & \boldmath{$88.18(0.12)$} & \boldmath{$87.63(0.25)$} & $87.04\,(0.36)$ & \boldmath{$86.39(0.45)$} & \boldmath{$85.59(0.47)$} & \boldmath{$84.80(0.33)$} & \boldmath{$82.93(0.40)$} & \boldmath{$79.13(0.27)$} \\
& PRODEN & $88.38\,(0.26)$ & $87.92\,(0.36)$ & $87.51\,(0.01)$ & \boldmath{$87.46(0.05)$} & $86.38\,(0.32)$ & $85.22\,(0.46)$ & $83.62\,(0.27)$ & $80.90\,(0.56)$ & $74.00\,(1.87)$ \\
& PC & $87.29\,(0.18)$ & $83.18\,(0.44)$ & $81.70\,(0.07)$ & $80.51\,(0.37)$ & $79.31\,(0.38)$ & $77.65\,(0.56)$ & $76.36\,(0.88)$ & $74.34\,(1.10)$ & $69.69\,(1.40)$ \\
& OVA & $83.56\,(0.69)$ & $77.59\,(2.11)$ & $77.86\,(1.90)$ & $73.61\,(2.87)$ & $63.45\,(7.17)$ & $66.08\,(3.39)$ & $59.62\,(9.95)$ & $62.20\,(4.80)$ & $55.03\,(7.38)$ \\

\midrule

\multirow{4}{*}{\rotatebox[origin=c]{90}{ Kuzushiji }} & \our{} & \boldmath{$80.81(0.88)$} & $79.49\,(0.58)$ & \boldmath{$78.36(0.89)$} & \boldmath{$75.83(0.53)$} & \boldmath{$73.68(0.39)$} & \boldmath{$71.57(0.46)$} & \boldmath{$68.42(0.52)$} & \boldmath{$65.75(0.35)$} & \boldmath{$56.39(0.36)$} \\
& PRODEN & $79.92\,(0.20)$ & \boldmath{$79.78(1.21)$} & $77.79\,(0.40)$ & $74.65\,(0.29)$ & $72.95\,(0.84)$ & $69.04\,(1.07)$ & $63.93\,(1.20)$ & $56.91\,(0.09)$ & $38.08\,(2.13)$ \\
& PC & $78.67\,(0.75)$ & $69.25\,(0.30)$ & $64.92\,(0.81)$ & $62.81\,(1.40)$ & $60.45\,(0.80)$ & $57.56\,(2.53)$ & $53.82\,(0.83)$ & $50.36\,(1.19)$ & $44.48\,(0.38)$ \\
& OVA & $75.91\,(1.03)$ & $70.40\,(1.12)$ & $65.82\,(0.49)$ & $62.53\,(1.12)$ & $57.41\,(0.85)$ & $55.11\,(0.34)$ & $46.93\,(2.19)$ & $44.65\,(3.07)$ & $34.43\,(0.64)$ \\

\midrule

\multirow{4}{*}{\rotatebox[origin=c]{90}{ CIFAR-10 }} & \our{} & \boldmath{$72.91(0.37)$} & $67.46\,(1.39)$ & $65.49\,(1.52)$ & $62.91\,(1.38)$ & $61.14\,(0.98)$ & \boldmath{$56.48(0.65)$} & \boldmath{$51.87(1.63)$} & \boldmath{$45.69(0.34)$} & \boldmath{$38.48(1.66)$} \\
& PRODEN & $66.94\,(3.99)$ & \boldmath{$69.10(1.35)$} & \boldmath{$66.83(0.66)$} & \boldmath{$64.13(0.75)$} & \boldmath{$61.26(0.81)$} & $56.28\,(1.41)$ & $49.51\,(1.15)$ & $37.67\,(0.15)$ & $23.45\,(2.10)$ \\
& PC & $70.62\,(0.42)$ & $59.22\,(0.40)$ & $54.26\,(0.64)$ & $50.23\,(0.41)$ & $45.82\,(0.42)$ & $43.32\,(1.25)$ & $39.92\,(0.34)$ & $35.85\,(0.75)$ & $30.90\,(0.47)$ \\
& OVA & $69.76\,(0.03)$ & $59.07\,(0.56)$ & $52.58\,(0.21)$ & $46.79\,(0.97)$ & $43.23\,(0.67)$ & $40.28\,(2.03)$ & $35.08\,(0.63)$ & $33.49\,(1.31)$ & $29.97\,(1.24)$ \\

\bottomrule
\end{tabular}
\caption{Test accuracy (mean\,(std)\%) for 10 classes and the increasing number of candidate labels. The highest accuracy is bolded. Our ProPaLL overpass baseline methods, especially for a larger number of candidate labels.
}
\label{tab.cll_k10}
\end{table*}

\begin{table}[h]\fontsize{8}{10}\selectfont
\centering
\begin{tabular}{ @{}l@{}l@{\,}c@{}c@{}c@{}c@{} }
\toprule
& \multirow{2}{*}{Methods} & \multicolumn{4}{c}{Number of candidate labels} \\
\cmidrule{3-6}
&  & 1 & 2 & 3 & 4 \\
\midrule
\multirow{4}{*}{\rotatebox[origin=c]{90}{ MNIST }} & \our{} & \boldmath{$98.83(0.13)$} & \boldmath{$98.57(0.20)$} & \boldmath{$98.27(0.11)$} & \boldmath{$97.47(0.18)$} \\
& PRODEN & $98.73\,(0.26)$ & $98.37\,(0.05)$ & $98.23\,(0.37)$ & $96.92\,(0.18)$ \\
& PC & $98.63\,(0.04)$ & $96.82\,(0.12)$ & $95.88\,(0.21)$ & $94.54\,(1.04)$ \\
& OVA & $98.55\,(0.19)$ & $97.81\,(0.27)$ & $96.62\,(0.29)$ & $94.31\,(0.27)$ \\

\midrule

\multirow{4}{*}{\rotatebox[origin=c]{90}{ Fashion }} & \our{} & $89.79\,(0.27)$ & $89.19\,(0.13)$ & $88.02\,(0.25)$ & \boldmath{$86.31(1.13)$} \\
& PRODEN & \boldmath{$89.92(0.18)$} & \boldmath{$89.27(0.27)$} & \boldmath{$88.45(0.27)$} & $85.85\,(0.47)$ \\
& PC & $88.75\,(0.22)$ & $86.14\,(0.06)$ & $84.42\,(1.09)$ & $81.28\,(0.96)$ \\
& OVA & $80.51\,(0.61)$ & $81.61\,(1.81)$ & $78.51\,(0.97)$ & $78.01\,(2.03)$ \\

\midrule

\multirow{4}{*}{\rotatebox[origin=c]{90}{ Kuzushiji }} & \our{} & \boldmath{$88.15(0.59)$} & \boldmath{$86.17(0.37)$} & \boldmath{$83.68(0.84)$} & \boldmath{$78.82(1.91)$} \\
& PRODEN & $87.59\,(0.88)$ & $85.81\,(0.64)$ & $81.93\,(1.31)$ & $76.44\,(0.30)$ \\
& PC & $86.13\,(0.52)$ & $79.91\,(1.12)$ & $76.07\,(0.36)$ & $70.00\,(1.74)$ \\
& OVA & $86.41\,(0.32)$ & $80.34\,(0.71)$ & $74.27\,(1.22)$ & $67.07\,(1.69)$ \\

\midrule

\multirow{4}{*}{\rotatebox[origin=c]{90}{ CIFAR-10 }} & \our{} & \boldmath{$78.98(1.06)$} & \boldmath{$75.03(2.43)$} & \boldmath{$70.94(0.66)$} & \boldmath{$62.25(0.20)$} \\
& PRODEN & $73.25\,(0.77)$ & $71.63\,(2.77)$ & $69.77\,(0.45)$ & $60.26\,(0.50)$ \\
& PC & $77.48\,(1.01)$ & $68.29\,(0.84)$ & $63.20\,(1.07)$ & $56.28\,(0.30)$ \\
& OVA & $77.71\,(0.10)$ & $68.51\,(0.84)$ & $62.32\,(0.79)$ & $54.31\,(1.37)$ \\
\bottomrule
\end{tabular}
\caption{Test accuracy (mean\,(std)\%) for 5 classes and the increasing number of candidate labels. The highest accuracy is bolded.
}
\label{tab.cll_k5}
\end{table}

\paragraph{Results on CLL datasets.}
In~\Cref{tab.cll_k10,tab.cll_k5}, we analyze how models behave if the size of candidate label set increases. For this purpose, we adapted the experiment from~\cite{katsura2020bridging} with MNIST, Fashion-MNIST, Kuzushiji-MNIST, and CIFAR-10 datasets, for $10$ or $5$ classes\footnote{Datasets with $5$ classes were created by extracting images belonging to labels from $5$ to $9$.}. The results obtained for $3$ repetitions of each experiment show that model precision decreases with the increasing number of candidate labels. Moreover, for $9$ (or $4$) candidate labels (the most challenging setup), our approach returns the highest accuracy for all datasets. The highest gain compared to the second-best method is obtained for the MNIST and Kuzushiji-MNIST databases ($14.96\%$ and $11.91\%$, respectively). However, a similar trend is also observed for the remaining datasets.

\begin{figure}[t]
    \centering
    \includegraphics[width=\columnwidth]{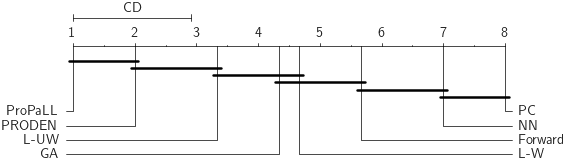}
    \caption{Critical difference diagrams comparing results of the considered methods shown in~\Cref{tab.cll} (smaller is better).}
    \label{fig:cd_cll}
\end{figure}

We also evaluate considered methods on unbiased complementary labels, as in~\cite{gao2021discriminative}. The results presented in~\Cref{tab.cll} for $10$ repetitions show that our approach achieves the highest accuracy. Moreover, as presented in the critical difference diagram in~\Cref{fig:cd_cll} it is statistically better than all baseline methods except PRODEN.

\begin{figure}[t]
    \centering
    \includegraphics[width=\columnwidth]{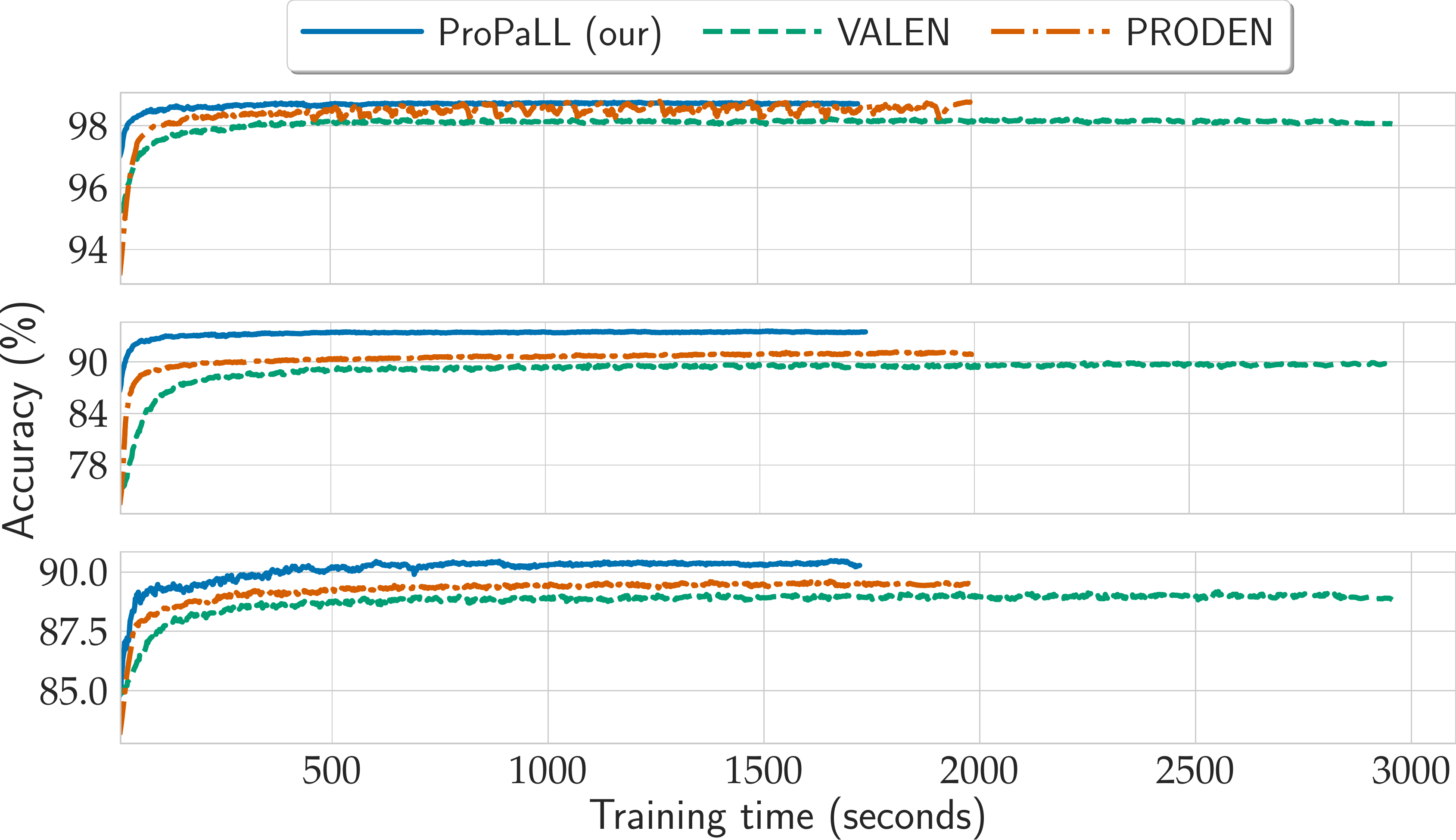}
    \caption{Training time of ProPaLL and two competitive methods on three datasets (MNIST, Kuzushiji-MNIST, Fashion-MNIST from top to bottom). Even though each model was trained for 500 epochs, \our{} worked faster than PRODEN and VALEN due to the simplicity of the implementation. Moreover, our model achieves accuracy close to the final accuracy of the baseline model already in the initial state of training.}
    \label{fig:relativeTime}
\end{figure}

\paragraph{Behavior of ProPall during training.}
Our cost function from~\Cref{eq:pll} is simple and requires only a few numerical operations, in contrast to the more complex baseline methods. Therefore, it is competitive in terms of numerical efficiency, as observed in~\Cref{fig:relativeTime}, where our method finishes $500$ learning epochs much faster than VALEN and PRODEN methods.
Moreover, it gradually learns consecutive classes instead of progressively increasing sensitivity for all of them. For example, as presented in~\Cref{fig:jump_acc}, a model learns most of the classes at the initial training stage and then continues training by considering the remaining classes one after another.

\paragraph{Gumbel noice influence.}
To provide a convincing argument for using Gumbel noise as a regularization technique, we run an ablation study for real-world databases that test the performance of ProPaLL with and without the noise. As shown in~\Cref{tab.noice_influence}, the experiment performed on the real-world datasets reveals that adding noise usually slightly increases the accuracy.

\begin{table}[t]\fontsize{9.8}{10.8}\selectfont 
\centering
\begin{tabular}{ @{}l@{\,}c@{\,}c@{\,}c@{} }
\toprule
\multirow{2}{*}{Methods} & \multicolumn{3}{c}{Datasets}\\
\cmidrule{2-4}
 & MNIST & Fashion & Kuzushiji \\
\midrule
\our{} (our) & \boldmath{$96.11(0.19)$} & \boldmath{$85.50(0.05)$} & \boldmath{$76.01(1.70)$}\\
PRODEN & $94.83\,(0.09)$ & $84.53\,(0.08)$ & $75.21\,(0.49)$ \\
PC & $84.04\,(0.55)$ & $77.55\,(0.39)$ & $59.32\,(0.59)$ \\
L-W & $92.38\,(0.19)$ & $83.82\,(0.29)$ & $66.86\,(1.95)$ \\
L-UW & $92.77\,(0.21)$ & $84.01\,(0.23)$ & $67.22\,(2.17)$ \\
Forward & $91.93\,(0.25)$ & $82.31\,(0.24)$ & $ 65.59\,(0.54)$ \\
NN & $89.99\,(0.42)$ & $ 80.29\,(0.47)$ & $65.44\,(0.51)$ \\
GA & $92.49\,(0.25)$ & $81.62\,(0.19)$ & $69.56\,(0.53)$ \\
\bottomrule
\end{tabular}
\caption{Test accuracy (mean\,(std)\%) of models trained on data with unbiased complementary labels. The highest accuracy is bolded.
}
\label{tab.cll}
\end{table}

\begin{table}[t]\fontsize{9.8}{10.8}\selectfont 
\centering
 \begin{tabular}{l@{\;\;}c@{\;\;}c@{}} 
 \toprule
 \multirow{2}{*}{Dataset} & \multicolumn{2}{c}{Methods} \\
 \cmidrule{2-3}
 & \our{} (w/o noise) & \our{} \\
 \midrule
 Lost & \boldmath{$79.00\,(0.91)$} &  $78.78\,(0.22)$ \\
 MSRCv2 & $41.02\,(1.57)$ & \boldmath{$42.05\,(0.58)$} \\
 BirdSong & $70.26\,(0.46)$ & \boldmath{$70.49\,(0.56)$} \\
 Soccer Player & $56.19\,(0.31)$ & \boldmath{$56.73\,(0.27)$} \\
 Yahoo!News & $67.86\,(0.09)$ & \boldmath{$68.06\,(0.08)$} \\
 \bottomrule
 \end{tabular}
\caption{Test accuracy (mean\,(std)\%) of our approach without and with Gumbel noise regularization on the real-world datasets. The highest accuracy is bolded. The noise usually increases model performance.}
\label{tab.noice_influence}
\end{table}

\section{Conclusions}

In this paper, we introduced \our{}, a simple and easy-to-implement probabilistic approach to partial label learning. We derived the formula for novel cost function starting from BCE and compared our model with state-of-the-art methods, like PRODEN and VALEN. The extensive experiments showed that \our{} usually outperforms existing methods and, at the same time, is trained up to two times faster. Moreover, a detailed training analysis indicated that it gradually learns consecutive classes instead of progressively increasing sensitivity for all of them.

In the future, we would like to concentrate on generalizing our function to more general setups, like multi-class classification, and adapt it to other target tasks, such as detection or segmentation.

\section*{Acknowledgements}
The work of \L. Struski was supported by the National Centre of Science (Poland) Grant No. 2020/39/D/ST6/01332\footnote{For the purpose of Open Access, the author has applied a CC-BY public copyright license to any Author Accepted Manuscript version arising from this submission.}.
J. Tabor and B. Zieli\'nski carried out this work within the research project ``Bio-inspired artificial neural network'' (grant no. POIR.04.04.00-00-14DE/18-00) within the Team-Net program of the Foundation for Polish Science co-financed by the European Union under the European Regional Development Fund.

\appendix
 
\counterwithin{figure}{section}
\counterwithin{table}{section}

\begin{appendices}
\crefalias{section}{appsec}

\section{Efficient Implementation of ProPaLL Loss Function}

Recall that the cost function for a point $x$ is given by
$$
\cost(x)=\underbrace{-\log \big(1-\prod_{i \in S} (1-p_i(x)) \big)}_{I} \underbrace{-\sum_{j \not\in S} \log(1-p_j(x))}_{II},
$$
where $S$ denotes the candidate label set of $x$. Moreover, let $S[\cdot]$ denotes the characteristic function of $S$. Since
$$
-\log(1-\sigma(r))=-\log (\tfrac{1}{1+\exp(r)})=\mathrm{LogSumExp}(0,r),
$$
part $II$ reduces to
$$
II=\sum_{j \not\in S} \mathrm{LogSumExp}(0,r_j(x))
$$
$$
=
\sum_{j} (1- S[j]) \cdot \mathrm{LogSumExp}(0,r_j(x)),
$$
and $I$ reduces to
$$
I=
-\log \big(1-\exp(-\sum_{i} S[i] \cdot \mathrm{LogSumExp}(0,r_i(x))
) \big).
$$
To stabilize $I$, let us observe that for small $h>0$
$$
-\log(1-\exp(-h))=-\log(h-\frac{1}{2!}h^2+\frac{1}{3!}h^3\pm..)
$$
$$
 =-\log(h)+\log(1-\frac{1}{2!}h+\frac{1}{3!}h^2\pm..)
=-\log(h)+O(h).
$$
Hence, when the sum inside logarithm is close to $1$ (i.e. when $r_i(x)$ are large negative numbers for $s \in S[i]$), then
$$
I\approx -\log (\sum_{i} S[i] \cdot \log (1+\exp(r_i(x)))) 
$$
$$
\approx
-\log (\sum_{i} S[i] \cdot \exp(r_i(x)))
$$
$$
=-\mathrm{LogSumExp}((S[i] \cdot r_i(x) +(1-S[i]) \cdot minF)_{i=0..k-1}),
$$
where $minF$ denotes the smallest numerical float (with $\exp(minF)=0$). Finally, we obtain
\[
I = -\log \Big(1-\exp\big(-\sum_{i} S[i] \cdot \mathrm{LogSumExp}(0,r_i(x)) \big) \Big)
\]
if $\max \limits_{i=0..k-1}\big(S[i] \cdot r_i(x) +(1-S[i]) \cdot minF\big) > -10$ and otherwise
\[
I = -\mathrm{LogSumExp}\Big(\big(S[i] \cdot r_i(x) +(1-S[i]) \cdot minF\big)_{i}\Big).
\]


\begin{figure*}[t!]
    \renewcommand\thefigure{D.1}
    \centering
    \includegraphics[width=\textwidth]{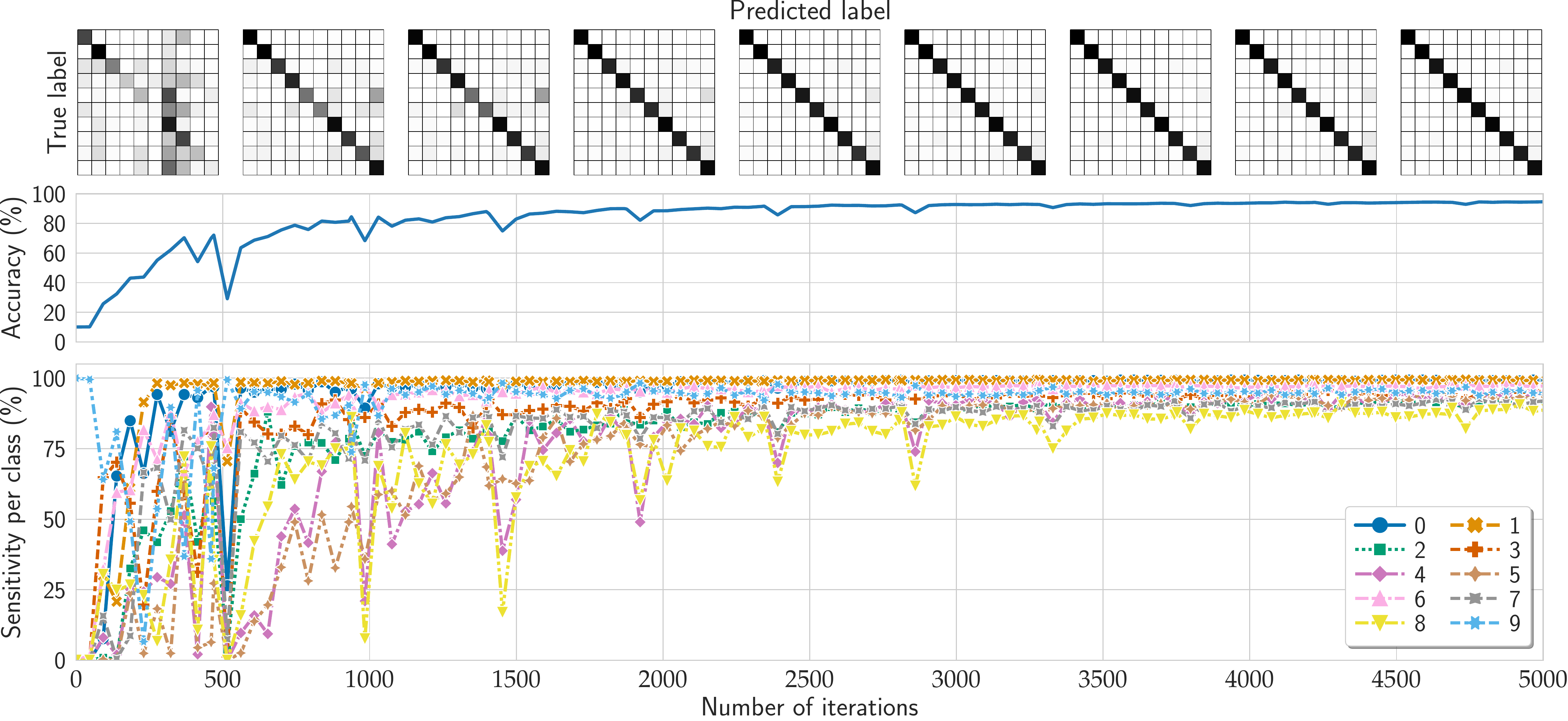}
    \caption{Detailed analysis of the PRODEN training (presented here) indicates that it differs from ProPaLL training (shown in~\Cref{fig:jump_acc}). While PRODEN progressively increases sensitivity for all classes, ProPaLL gradually learns consecutive classes. In this image, the middle part presents model accuracy for all classes in time, the bottom part shows the sensitivity of particular classes also in time, while the top part shows confusion matrices in time points corresponding to their location.}
    \label{fig:jump_acc_proden}
\end{figure*}

\section{Derivation of Gumbel Noise}

In the standard softmax model, noise from the Gumbel distributions is often added to the logits to widen the space of parameters search. More precisely, logits $r_1,\ldots,r_n$ are transformed to
$$
r_1+\lambda U_1,\ldots,r_n+ \lambda U_n,
$$
where $U_i$ are independently sampled
from Gumbel distribution, and $\lambda$ is a parameter (usually set to $1$ at the beginning of training, which is decreased to $0$ during training). Then, for two classes
$$
s_0=r_0+\lambda U_0, s_1=r_1+\lambda U_1.
$$
and after applying softmax
$$
p_0=\frac{\exp(s_0)}{\exp(s_0)+\exp(s_1)}=\frac{1}{1+\exp(s_1-s_0)},
$$
$$
p_1=\frac{1}{1+\exp(s_0-s_1)}. 
$$
Thus
$$
p_1=\sigma(s_1-s_0) \text{ and }
p_0=\sigma(-(s_1-s_0)).
$$
On the other hand, in the case of binary cross-entropy, where the representation corresponds to one parameter $r=r_1-r_0$, adding Gumbel noise $\lambda U$ to the softmax logits is equivalent to
$$
r = r + \lambda (U-V),
$$
where $U$ and $V$ are independent and come from the Gumbel distribution. Since ProPaLL uses binary cross entropy with $k$ outputs, we add this class to the logit of each class.

\section{Experimental Setup Details}
For PLL setting, we base on experiments and databases described in~\cite{lv2020progressive,xu2021instance} and the corresponding source codes available at \url{https://github.com/palm-ml/valen} and \url{https://github.com/Lvcrezia77/PRODEN}.
For corrupted MNIST, Fashion-MNIST, and Kuzushiji-MNIST datasets, we use MLP with $5$ fully-connected layers (with 300, 301, 302, 303, and 10 neurons), ReLU as the activation function, and batch normalization~\cite{ioffe2015batch} applied before hidden layers. For random corruption, this model is trained with batch size 256, learning rate 0.05, and weight decay $10^{-6}$. For instance-dependent corruption, we use batch size $256$, learning rate $0.1$ or $0.05$, and weight decay $0$.
For both corruption types of CIFAR-10 database, we use ResNet-32~\cite{he2016deep} with batch size $256$, learning rate $0.05$, and weight decay $0$. 
For the real-world PLL datasets, we use linear model with learning rate $0.01$, $0.05$, $0.1$, $0.5$, $0.5$ for Lost, MSRCv2, BirdSong, Soccer Player, and Yahoo!News, respectively. Weight decay equals $10^{-4}$ for Lost and MSRCv2, and $0$ for the remaining datasets. Mini-batch is set to $100$ for Lost, MSRCv2, and BirdSong, and $256$ for Soccer Player and Yahoo!News.

\begin{table*}[t]\fontsize{8}{10}\selectfont
\renewcommand\thetable{D.1}
\renewcommand{\arraystretch}{1.5}
\centering
\begin{tabular}{@{}l@{}l@{}c@{\,}c@{\,}c@{\,}c@{\,}c@{\,}c@{\,}c@{\,}c@{\,}c@{}}
\toprule
& \multirow{2}{*}{Methods} & \multicolumn{9}{c}{Number of candidate labels} \\
\cmidrule(lr){3-11}
& & 1 & 2 & 3 & 4 & 5 & 6 & 7 & 8 & 9 \\
\midrule

\multirow{2}{*}{\rotatebox[origin=l]{90}{MNIST}} & \makecell[l]{\our{} \\ (w/o noise)} & \boldmath{$96.44(0.09)$} & $95.99\,(0.18)$ & \boldmath{$95.89(0.19)$} & \boldmath{$95.47(0.10)$} & \boldmath{$95.19(0.07)$} & \boldmath{$94.77(0.43)$} & $93.80\,(0.21)$ & $92.33\,(0.29)$ & $89.77\,(0.69)$ \\
& \our{} & $96.28\,(0.08)$ & \boldmath{$95.99(0.07)$} & $95.83\,(0.28)$ & $95.45\,(0.22)$ & $95.10\,(0.11)$ & $94.62\,(0.08)$ & \boldmath{$93.86(0.22)$} & \boldmath{$92.56(0.27)$} & \boldmath{$90.17(0.24)$} \\

\midrule

\multirow{2}{*}{\rotatebox[origin=l]{90}{ Fashion }} & \makecell[l]{\our{} \\ (w/o noise)} & $88.44\,(0.35)$ & $88.12\,(0.34)$ & \boldmath{$87.74(0.63)$} & \boldmath{$87.50(0.42)$} & \boldmath{$86.59(0.28)$} & \boldmath{$85.64(0.24)$} & $84.49\,(0.11)$ & \boldmath{$83.10(0.15)$} & \boldmath{$79.28(0.33)$} \\
 & \our{} & \boldmath{$88.73(0.25)$} & \boldmath{$88.18(0.12)$} & \boldmath{$87.63(0.25)$} & $87.04\,(0.36)$ & $86.39\,(0.45)$ & $85.59\,(0.47)$ & \boldmath{$84.80(0.33)$} & $82.93\,(0.40)$ & $79.13\,(0.27)$ \\

\midrule

\multirow{2}{*}{\rotatebox[origin=l]{90}{ \parbox[c]{1.05cm}{\centering Kuzushiji} }} & \makecell[l]{\our{} \\ (w/o noise)} & \boldmath{$80.91\,(0.33)$} & \boldmath{$79.67(1.04)$} & \boldmath{$78.44(0.10)$} & $75.10\,(0.46)$ & $73.13\,(0.61)$ & $69.79\,(1.46)$ & $67.01\,(1.64)$ & $64.31\,(1.56)$ & $55.67\,(3.24)$ \\
& \our{} & $80.81\,(0.88)$ & $79.49\,(0.58)$ & $78.36\,(0.89)$ & \boldmath{$75.83(0.53)$} & \boldmath{$73.68(0.39)$} & \boldmath{$71.57(0.46)$} & \boldmath{$68.42(0.52)$} & \boldmath{$65.75(0.35)$} & \boldmath{$56.39(0.36)$} \\

\midrule

\multirow{2}{*}{\rotatebox[origin=l]{90}{ \parbox[c]{1.1cm}{\centering CIFAR10} }} & \makecell[l]{\our{} \\ (w/o noise)} & $71.47\,(1.64)$ & \boldmath{$69.85(1.08)$} & \boldmath{$67.51(1.13)$} & \boldmath{$64.91(1.60)$} & \boldmath{$62.47(0.84)$} & \boldmath{$56.97(1.02)$} & \boldmath{$54.23(1.12)$} & \boldmath{$48.05(1.37)$} & \boldmath{$39.95(0.98)$} \\
& \our{} & \boldmath{$72.91(0.37)$} & $67.46\,(1.39)$ & $65.49\,(1.52)$ & $62.91\,(1.38)$ & $61.14\,(0.98)$ & $56.48\,(0.65)$ & $51.87\,(1.63)$ & $45.69\,(0.34)$ & $38.48\,(1.66)$ \\

\bottomrule
\end{tabular}
\caption{Test accuracy (mean\,(std)\%) of ProPaLL with and without Gumbel noise for 10 classes and the increasing number of candidate labels. The highest accuracy is bolded.
}
\label{tab.cll_k10_app}
\end{table*}

\begin{table}[t]\fontsize{9.8}{10.8}\selectfont
\renewcommand\thetable{D.2}
\centering
 \begin{tabular}{l@{\,}c@{\;}c@{} }
 \toprule
 \multirow{2}{*}{Datasets} & \multicolumn{2}{c}{Methods} \\
 \cmidrule{2-3}
  & \our{} (w/o noise) & \our{} \\
  \midrule
  MNIST & $98.68\,(0.03)$ & \boldmath{$98.73\,(0.05)$} \\
  Kuzushiji & $93.30\,(0.21)$ & \boldmath{$93.50\,(0.17)$} \\
  Fashion & $90.40\,(0.23)$ & \boldmath{$90.40\,(0.06)$} \\
  CIFAR-10 & $83.90\,(0.22)$ & \boldmath{$84.32\,(0.37)$} \\
 \bottomrule
 \end{tabular}
 \caption{Test accuracy (mean\,(std)\%) of ProPaLL with and without Gumbel noise on artificial datasets corrupted by the uniform generating procedure. The highest accuracy is bolded.
}
\label{tab.bemchmark_random_app}
\end{table}

\begin{table}[t]\fontsize{9.8}{10.8}\selectfont
\renewcommand\thetable{D.3}
\centering
 \begin{tabular}{l@{\,}c@{\;}c@{} }
 \toprule
 \multirow{2}{*}{Datasets} & \multicolumn{2}{c}{Methods} \\
 \cmidrule{2-3}
  & \our{} (w/o noise) & \our{} \\
  \midrule
  MNIST & $97.85\,(0.10)$ & \boldmath{$97.89(0.08)$} \\
  Kuzushiji & $86.14\,(0.37)$ & \boldmath{$86.26(0.32)$} \\
  Fashion & \boldmath{$85.63\,(0.09)$} & $85.55\,(0.33)$ \\
  CIFAR-10 & $74.13\,(1.33)$ & \boldmath{$75.75\,(0.63)$} \\ 
 \bottomrule
 \end{tabular}
 \caption{Test accuracy (mean\,(std)\%) of ProPaLL with and without Gumbel noise on artificial datasets corrupted by the instance-dependent generating procedure~\cite{xu2021instance}. The highest accuracy is bolded.
}
\label{tab.bemchmark_instance_dependent_app}
\end{table}

For CLL setting, we base on experiments and databases described in~\cite{katsura2020bridging,gao2021discriminative} and the corresponding source codes available at \url{https://github.com/YasuhiroKatsura/ord-comp} and \url{https://github.com/GaoYi439/complementary-label-learning}.
We adapted the experiment with MNIST, Fashion-MNIST, Kuzushiji-MNIST, and CIFAR-10 datasets with $10$ or $5$ classes. For MNIST, Fashion-MNIST, and Kuzushiji-MNIST datasets, we use MLP with $2$ fully-connected layers (with 500 and 10 neurons), ReLU as the activation function, batch size 64, weight decay $10^{-4}$, and learning rate $5\cdot 10^{-4}$. For CIFAR-10, we apply DenseNet~\cite{huang2017densely} with weight decay $5\cdot 10^{-4}$, learning rate $0.01$, and momentum $0.9$.
For unbiased complementary labels, as in~\cite{gao2021discriminative}, we use the same MLP model, Adam optimization, number of epoch $300$, and mini-batch $256$. Moreover, for the MNIST and Kuzushiji-MNIST datasets, we use learning rate $10^{-3}$ and weight decay $10^{-5}$. For Fashion-MNIST, we used learning rate $0.001$ and  weight decay $0$.

\section{Additional Results}

\setcounter{table}{3} 

\paragraph{Behavior of PRODEN during training.}
To highlight the difference in ProPaLL training (shown in~\Cref{fig:jump_acc}) compared to existing methods, we provide similar details on PRODEN training in~\Cref{fig:jump_acc_proden}. One can observe that, while ProPaLL gradually learns consecutive classes, PRODEN progressively increases sensitivity for all of them. It can be caused by  EM~\cite{jin2002learning} inspirations of PRODEN and the fact that ``frequent patterns'' are remembered for all classes already from the initial training iterations thanks to the memorization effects~\cite{arpit2017closer,han2018co}.


\paragraph{Gumbel noice influence.}
Finally, we provide test accuracy of our approach without and with Gumbel noise regularization for the remaining settings and datasets (complementary to~\Cref{tab.noice_influence}). In~\Cref{tab.bemchmark_random_app,tab.bemchmark_instance_dependent_app}, we report results for artificial databases corrupted by the uniform and instance-dependent generating procedure, respectively. In~\Cref{tab.cll_k10_app,tab.cll_k5_app}, we analyze how \our{} with and without noise behave if the size of candidate label set increases.
Finally, in~\Cref{tab.cll_app}, we provide results for unbiased complementary labels.
We conclude that adding noise to the logits of the model usually improves the results.

\begin{table}[h]\fontsize{9.8}{10.8}\selectfont 
\centering
\begin{tabular}{l@{\,}c@{\;}c@{} }
 \toprule
 \multirow{2}{*}{Datasets} & \multicolumn{2}{c}{Methods} \\
 \cmidrule{2-3}
  & \our{} (w/o noise) & \our{} \\
  \midrule
 MNIST & $96.10\,(0.14)$ & \boldmath{$96.11(0.19)$} \\
 Fashion & $85.43\,(0.03)$ & \boldmath{$85.50(0.05)$} \\
 Kuzushiji & $72.61\,(0.63)$ & \boldmath{$76.01(1.70)$}\\
\bottomrule
\end{tabular}
\caption{Test accuracy (mean\,(std)\%)  of ProPaLL with and without Gumbel noise trained on data with unbiased complementary labels. The highest accuracy is bolded.
}
\label{tab.cll_app}
\end{table}

\begin{table}[h]\fontsize{9}{10}\selectfont
\centering
\begin{tabular}{ @{}l@{\quad}c@{\quad}l@{\;\;}c@{\;}c@{} }
\cmidrule{2-5}
 & & \multirow{2}{*}{Datasets} & \multicolumn{2}{c}{Methods} \\
\cmidrule(r){4-5}
 & & & \our{} (w/o noise) & \our{} \\
\midrule
\multirow{17}{*}{\rotatebox[origin=c]{90}{Number of candidate labels}} 
& \multirow{4}{*}{1} & MNIST & $98.80\,(0.06)$ & \boldmath{$98.83(0.13)$} \\
& & Fashion & \boldmath{$90.45\,(0.79)$} & $89.79\,(0.27)$ \\
& & Kuzushiji & $87.84\,(0.49)$ & \boldmath{$88.15(0.59)$} \\
& & CIFAR-10 & $78.39\,(0.66)$ & \boldmath{$78.98(1.06)$} \\
\cmidrule{2-5}
& \multirow{4}{*}{2} & MNIST & \boldmath{$98.69\,(0.11)$} & $98.57(0.20)$ \\
& & Fashion & \boldmath{$89.27\,(0.64)$} & $89.19\,(0.13)$ \\
& & Kuzushiji & \boldmath{$86.55\,(0.34)$} & \boldmath{$86.17(0.37)$} \\
& & CIFAR-10 & $74.73\,(1.49)$ & \boldmath{$75.03(2.43)$} \\
\cmidrule{2-5}
& \multirow{4}{*}{3} & MNIST & $98.24\,(0.19)$ & \boldmath{$98.27(0.11)$} \\
& & Fashion & \boldmath{$88.30\,(0.20)$} & $88.02\,(0.25)$ \\
& & Kuzushiji & $83.13\,(1.42)$ & \boldmath{$83.68(0.84)$} \\
& & CIFAR-10 & $69.51\,(1.80)$ & \boldmath{$70.94(0.66)$} \\
\cmidrule{2-5}
& \multirow{4}{*}{4} & MNIST & \boldmath{$97.83\,(0.34)$} & $97.47(0.18)$ \\
& & Fashion & $86.18\,(0.45)$ & \boldmath{$86.31(1.13)$} \\
& & Kuzushiji & $78.16\,(1.43)$ & \boldmath{$78.82(1.91)$} \\
& & CIFAR-10 & \boldmath{$62.45\,(0.48)$} & $62.25(0.20)$ \\
\bottomrule
\end{tabular}
\caption{Test accuracy (mean\,(std)\%) of ProPaLL with and without Gumbel noise for 5 classes and the increasing number of candidate labels. The highest accuracy is bolded.
}
\label{tab.cll_k5_app}
\end{table}

\end{appendices}


\bibliography{refs}

\begin{thebibliography}{43}
\providecommand{\natexlab}[1]{#1}
\providecommand{\url}[1]{\texttt{#1}}
\expandafter\ifx\csname urlstyle\endcsname\relax
  \providecommand{\doi}[1]{doi: #1}\else
  \providecommand{\doi}{doi: \begingroup \urlstyle{rm}\Url}\fi

\bibitem[Armato~III et~al.(2011)Armato~III, McLennan, Bidaut, McNitt-Gray,
  Meyer, Reeves, Zhao, Aberle, Henschke, Hoffman, et~al.]{armato2011lung}
Armato~III, S.~G., McLennan, G., Bidaut, L., McNitt-Gray, M.~F., Meyer, C.~R.,
  Reeves, A.~P., Zhao, B., Aberle, D.~R., Henschke, C.~I., Hoffman, E.~A.,
  et~al.
\newblock The lung image database consortium (lidc) and image database resource
  initiative (idri): a completed reference database of lung nodules on ct
  scans.
\newblock \emph{Medical physics}, 38\penalty0 (2):\penalty0 915--931, 2011.

\bibitem[Arpit et~al.(2017)Arpit, Jastrzebski, Ballas, Krueger, Bengio, Kanwal,
  Maharaj, Fischer, Courville, Bengio, et~al.]{arpit2017closer}
Arpit, D., Jastrzebski, S., Ballas, N., Krueger, D., Bengio, E., Kanwal, M.~S.,
  Maharaj, T., Fischer, A., Courville, A., Bengio, Y., et~al.
\newblock A closer look at memorization in deep networks.
\newblock In \emph{International conference on machine learning}, pp.\
  233--242. PMLR, 2017.

\bibitem[Briggs et~al.(2012)Briggs, Fern, and Raich]{briggs2012rank}
Briggs, F., Fern, X.~Z., and Raich, R.
\newblock Rank-loss support instance machines for miml instance annotation.
\newblock In \emph{International Conference on Knowledge Discovery and Data
  Mining (SIGKDD)}, pp.\  534--542, 2012.

\bibitem[Chen et~al.(2017)Chen, Patel, and Chellappa]{chen2017learning}
Chen, C.-H., Patel, V.~M., and Chellappa, R.
\newblock Learning from ambiguously labeled face images.
\newblock \emph{Transactions on Pattern Analysis and Machine Intelligence},
  40\penalty0 (7):\penalty0 1653--1667, 2017.

\bibitem[Clanuwat et~al.(2018)Clanuwat, Bober-Irizar, Kitamoto, Lamb, Yamamoto,
  and Ha]{clanuwat2018deep}
Clanuwat, T., Bober-Irizar, M., Kitamoto, A., Lamb, A., Yamamoto, K., and Ha,
  D.
\newblock Deep learning for classical japanese literature.
\newblock \emph{arXiv preprint arXiv:1812.01718}, 2018.

\bibitem[Cour et~al.(2011)Cour, Sapp, and Taskar]{cour2011learning}
Cour, T., Sapp, B., and Taskar, B.
\newblock Learning from partial labels.
\newblock \emph{The Journal of Machine Learning Research}, 12:\penalty0
  1501--1536, 2011.

\bibitem[Dem{\v{s}}ar(2006)]{demvsar2006statistical}
Dem{\v{s}}ar, J.
\newblock Statistical comparisons of classifiers over multiple data sets.
\newblock \emph{The Journal of Machine learning research}, 7:\penalty0 1--30,
  2006.

\bibitem[Feng \& An(2019)Feng and An]{feng2019partial}
Feng, L. and An, B.
\newblock Partial label learning with self-guided retraining.
\newblock In \emph{AAAI conference on artificial intelligence}, volume~33, pp.\
   3542--3549, 2019.

\bibitem[Feng et~al.(2020)Feng, Lv, Han, Xu, Niu, Geng, An, and
  Sugiyama]{feng2020provably}
Feng, L., Lv, J., Han, B., Xu, M., Niu, G., Geng, X., An, B., and Sugiyama, M.
\newblock Provably consistent partial-label learning.
\newblock \emph{Advances in Neural Information Processing Systems (NeurIPS)},
  33:\penalty0 10948--10960, 2020.

\bibitem[Gao \& Zhang(2021)Gao and Zhang]{gao2021discriminative}
Gao, Y. and Zhang, M.-L.
\newblock Discriminative complementary-label learning with weighted loss.
\newblock In \emph{International Conference on Machine Learning (ICML)}, pp.\
  3587--3597, 2021.

\bibitem[Goodfellow et~al.(2014)Goodfellow, Pouget-Abadie, Mirza, Xu,
  Warde-Farley, Ozair, Courville, and Bengio]{goodfellow2014generative}
Goodfellow, I., Pouget-Abadie, J., Mirza, M., Xu, B., Warde-Farley, D., Ozair,
  S., Courville, A., and Bengio, Y.
\newblock Generative adversarial nets.
\newblock \emph{Advances in neural information processing systems (NeurIPS)},
  27, 2014.

\bibitem[Guillaumin et~al.(2010)Guillaumin, Verbeek, and
  Schmid]{guillaumin2010multiple}
Guillaumin, M., Verbeek, J., and Schmid, C.
\newblock Multiple instance metric learning from automatically labeled bags of
  faces.
\newblock In \emph{European conference on computer vision}, pp.\  634--647.
  Springer, 2010.

\bibitem[Han et~al.(2018)Han, Yao, Yu, Niu, Xu, Hu, Tsang, and
  Sugiyama]{han2018co}
Han, B., Yao, Q., Yu, X., Niu, G., Xu, M., Hu, W., Tsang, I., and Sugiyama, M.
\newblock Co-teaching: Robust training of deep neural networks with extremely
  noisy labels.
\newblock \emph{Advances in neural information processing systems}, 31, 2018.

\bibitem[He et~al.(2016)He, Zhang, Ren, and Sun]{he2016deep}
He, K., Zhang, X., Ren, S., and Sun, J.
\newblock Deep residual learning for image recognition.
\newblock In \emph{Conference on Computer Vision and Pattern Recognition
  (CVPR)}, pp.\  770--778, 2016.

\bibitem[Huang et~al.(2017)Huang, Liu, Van Der~Maaten, and
  Weinberger]{huang2017densely}
Huang, G., Liu, Z., Van Der~Maaten, L., and Weinberger, K.~Q.
\newblock Densely connected convolutional networks.
\newblock In \emph{Conference on Computer Vision and Pattern Recognition
  (CVPR)}, pp.\  4700--4708, 2017.

\bibitem[H{\"u}llermeier \& Beringer(2006)H{\"u}llermeier and
  Beringer]{hullermeier2006learning}
H{\"u}llermeier, E. and Beringer, J.
\newblock Learning from ambiguously labeled examples.
\newblock \emph{Intelligent Data Analysis}, 10\penalty0 (5):\penalty0 419--439,
  2006.

\bibitem[Ioffe \& Szegedy(2015)Ioffe and Szegedy]{ioffe2015batch}
Ioffe, S. and Szegedy, C.
\newblock Batch normalization: Accelerating deep network training by reducing
  internal covariate shift.
\newblock In \emph{International conference on machine learning}, pp.\
  448--456. PMLR, 2015.

\bibitem[Ishida et~al.(2017)Ishida, Niu, Hu, and Sugiyama]{ishida2017learning}
Ishida, T., Niu, G., Hu, W., and Sugiyama, M.
\newblock Learning from complementary labels.
\newblock \emph{Advances in Neural Information Processing Systems (NeurIPS)},
  30, 2017.

\bibitem[Ishida et~al.(2019)Ishida, Niu, Menon, and
  Sugiyama]{ishida2019complementary}
Ishida, T., Niu, G., Menon, A., and Sugiyama, M.
\newblock Complementary-label learning for arbitrary losses and models.
\newblock In \emph{International Conference on Machine Learning (ICML)}, pp.\
  2971--2980, 2019.

\bibitem[Jin \& Ghahramani(2002)Jin and Ghahramani]{jin2002learning}
Jin, R. and Ghahramani, Z.
\newblock Learning with multiple labels.
\newblock \emph{Advances in Neural Information Processing Systems}, 15, 2002.

\bibitem[Katsura \& Uchida(2020)Katsura and Uchida]{katsura2020bridging}
Katsura, Y. and Uchida, M.
\newblock Bridging ordinary-label learning and complementary-label learning.
\newblock In \emph{Asian Conference on Machine Learning (ACML)}, pp.\
  161--176, 2020.

\bibitem[Krizhevsky et~al.(2009)Krizhevsky, Hinton,
  et~al.]{krizhevsky2009learning}
Krizhevsky, A., Hinton, G., et~al.
\newblock Learning multiple layers of features from tiny images.
\newblock 2009.

\bibitem[LeCun et~al.(1998)LeCun, Bottou, Bengio, and
  Haffner]{lecun1998gradient}
LeCun, Y., Bottou, L., Bengio, Y., and Haffner, P.
\newblock Gradient-based learning applied to document recognition.
\newblock \emph{Proceedings of the IEEE}, 86\penalty0 (11):\penalty0
  2278--2324, 1998.

\bibitem[Liu \& Dietterich(2012)Liu and Dietterich]{liu2012conditional}
Liu, L. and Dietterich, T.
\newblock A conditional multinomial mixture model for superset label learning.
\newblock \emph{Advances in Neural Information Processing Systems}, 25, 2012.

\bibitem[Luo \& Orabona(2010)Luo and Orabona]{luo2010learning}
Luo, J. and Orabona, F.
\newblock Learning from candidate labeling sets.
\newblock \emph{Advances in Neural Information Processing Systems (NeurIPS)},
  23, 2010.

\bibitem[Lv et~al.(2020)Lv, Xu, Feng, Niu, Geng, and
  Sugiyama]{lv2020progressive}
Lv, J., Xu, M., Feng, L., Niu, G., Geng, X., and Sugiyama, M.
\newblock Progressive identification of true labels for partial-label learning.
\newblock In \emph{International Conference on Machine Learning (ICML)}, pp.\
  6500--6510, 2020.

\bibitem[Nguyen \& Caruana(2008)Nguyen and Caruana]{nguyen2008classification}
Nguyen, N. and Caruana, R.
\newblock Classification with partial labels.
\newblock In \emph{International Conference on Knowledge Discovery and Data
  Mining (SIGKDD)}, pp.\  551--559, 2008.

\bibitem[Patrini et~al.(2017)Patrini, Rozza, Krishna~Menon, Nock, and
  Qu]{patrini2017making}
Patrini, G., Rozza, A., Krishna~Menon, A., Nock, R., and Qu, L.
\newblock Making deep neural networks robust to label noise: A loss correction
  approach.
\newblock In \emph{Conference on Computer Vision and Pattern Recognition
  (CVPR)}, pp.\  1944--1952, 2017.

\bibitem[Struski et~al.(2021)Struski, Danel, {\'S}mieja, Tabor, and
  Zieli{\'n}ski]{struski2021song}
Struski, {\L}., Danel, T., {\'S}mieja, M., Tabor, J., and Zieli{\'n}ski, B.
\newblock Song: Self-organizing neural graphs.
\newblock \emph{arXiv preprint arXiv:2107.13214}, 2021.

\bibitem[Tang \& Zhang(2017)Tang and Zhang]{tang2017confidence}
Tang, C.-Z. and Zhang, M.-L.
\newblock Confidence-rated discriminative partial label learning.
\newblock In \emph{AAAI Conference on Artificial Intelligence}, volume~31,
  2017.

\bibitem[Wang et~al.(2021)Wang, Shangguan, Yang, Chuang, Zhou, Li, Venkatesh,
  Kalinli, and Chandra]{wang2021noisy}
Wang, D., Shangguan, Y., Yang, H., Chuang, P., Zhou, J., Li, M., Venkatesh, G.,
  Kalinli, O., and Chandra, V.
\newblock Noisy training improves e2e asr for the edge.
\newblock \emph{arXiv preprint arXiv:2107.04677}, 2021.

\bibitem[Xiao et~al.(2017)Xiao, Rasul, and Vollgraf]{xiao2017fashion}
Xiao, H., Rasul, K., and Vollgraf, R.
\newblock Fashion-mnist: a novel image dataset for benchmarking machine
  learning algorithms.
\newblock \emph{arXiv preprint arXiv:1708.07747}, 2017.

\bibitem[Xu et~al.(2019)Xu, Lv, and Geng]{xu2019partial}
Xu, N., Lv, J., and Geng, X.
\newblock Partial label learning via label enhancement.
\newblock In \emph{AAAI Conference on Artificial Intelligence}, volume~33, pp.\
   5557--5564, 2019.

\bibitem[Xu et~al.(2021)Xu, Qiao, Geng, and Zhang]{xu2021instance}
Xu, N., Qiao, C., Geng, X., and Zhang, M.-L.
\newblock Instance-dependent partial label learning.
\newblock \emph{Advances in Neural Information Processing Systems (NeurIPS)},
  34:\penalty0 27119--27130, 2021.

\bibitem[Xu et~al.(2020)Xu, Gong, Chen, Liu, Zhang, and
  Batmanghelich]{xu2020generative}
Xu, Y., Gong, M., Chen, J., Liu, T., Zhang, K., and Batmanghelich, K.
\newblock Generative-discriminative complementary learning.
\newblock In \emph{AAAI Conference on Artificial Intelligence}, volume~34, pp.\
   6526--6533, 2020.

\bibitem[Yao et~al.(2020)Yao, Deng, Chen, Gong, Wu, and Yang]{yao2020deep}
Yao, Y., Deng, J., Chen, X., Gong, C., Wu, J., and Yang, J.
\newblock Deep discriminative cnn with temporal ensembling for
  ambiguously-labeled image classification.
\newblock In \emph{AAAI Conference on Artificial Intelligence}, volume~34, pp.\
   12669--12676, 2020.

\bibitem[Yu \& Zhang(2016)Yu and Zhang]{yu2016maximum}
Yu, F. and Zhang, M.-L.
\newblock Maximum margin partial label learning.
\newblock In \emph{Asian Conference on Machine Learning (ACML)}, pp.\  96--111,
  2016.

\bibitem[Yu et~al.(2018)Yu, Liu, Gong, and Tao]{yu2018learning}
Yu, X., Liu, T., Gong, M., and Tao, D.
\newblock Learning with biased complementary labels.
\newblock In \emph{European Conference on Computer Vision (ECCV)}, pp.\
  68--83, 2018.

\bibitem[Zeng et~al.(2013)Zeng, Xiao, Jia, Chan, Gao, Xu, and
  Ma]{zeng2013learning}
Zeng, Z., Xiao, S., Jia, K., Chan, T.-H., Gao, S., Xu, D., and Ma, Y.
\newblock Learning by associating ambiguously labeled images.
\newblock In \emph{Conference on Computer Vision and Pattern Recognition
  (CVPR)}, pp.\  708--715, 2013.

\bibitem[Zhang \& Yu(2015)Zhang and Yu]{zhang2015solving}
Zhang, M.-L. and Yu, F.
\newblock Solving the partial label learning problem: An instance-based
  approach.
\newblock In \emph{International Joint Conference on Artificial Intelligence
  (IJCAI)}, 2015.

\bibitem[Zhang et~al.(2016)Zhang, Zhou, and Liu]{zhang2016partial}
Zhang, M.-L., Zhou, B.-B., and Liu, X.-Y.
\newblock Partial label learning via feature-aware disambiguation.
\newblock In \emph{International Conference on Knowledge Discovery and Data
  Mining (SIGKDD)}, pp.\  1335--1344, 2016.

\bibitem[Zhang et~al.(2017)Zhang, Yu, and Tang]{zhang2017disambiguation}
Zhang, M.-L., Yu, F., and Tang, C.-Z.
\newblock Disambiguation-free partial label learning.
\newblock \emph{IEEE Transactions on Knowledge and Data Engineering},
  29\penalty0 (10):\penalty0 2155--2167, 2017.

\bibitem[Zhou(2018)]{zhou2018brief}
Zhou, Z.-H.
\newblock A brief introduction to weakly supervised learning.
\newblock \emph{National Science Review}, 5\penalty0 (1):\penalty0 44--53,
  2018.

\end{thebibliography}
\bibliographystyle{icml2021}

\end{document}